\documentclass[runningheads]{llncs}

\usepackage{eccv}

\usepackage{eccvabbrv}

\usepackage{graphicx}
\usepackage{booktabs}
\usepackage{multirow}
\usepackage{arydshln}
\usepackage{marvosym}
\makeatletter
\def\blfootnote{\xdef\@thefnmark{}\@footnotetext}
\makeatother

\usepackage[accsupp]{axessibility}  

\usepackage{hyperref}

\usepackage{orcidlink}

\begin{document}

\title{DG-PIC: Domain Generalized Point-In-Context Learning for Point Cloud Understanding} 

\titlerunning{DG-PIC}

\author{Jincen Jiang\inst{1}$^*$\orcidlink{0000-0002-0150-4644} \and
Qianyu Zhou\inst{2}$^*$\orcidlink{0000-0002-5331-050X} \and
Yuhang Li\inst{1,3}\orcidlink{0000-0002-3827-1522} \and
Xuequan Lu\inst{4}\textsuperscript{\Letter}\orcidlink{0000-0003-0959-408X} \and\\
Meili Wang\inst{5}\textsuperscript{\Letter}\orcidlink{0000-0001-7901-1789} \and
Lizhuang Ma\inst{2}\orcidlink{0000-0003-1653-4341} \and
Jian Chang\inst{1}\orcidlink{0000-0003-4118-147X} \and
Jian Jun Zhang\inst{1}\orcidlink{0000-0002-7069-5771}
}

\authorrunning{J.~Jiang et al.}

\institute{National Centre for Computer Animation, Bournemouth University, Dorset, UK \and
Shanghai Jiao Tong University, Shanghai, China \and
Shanghai University, Shanghai, China \and
La Trobe University, Victoria, Australia \and
Northwest A\&F University, Yangling, China\\
\email{\{jiangj, jchang, jzhang\}@bournemouth.ac.uk}\\
\email{\{zhouqianyu, lzma\}@sjtu.edu.cn, yuhangli@shu.edu.cn}\\
\email{b.lu@latrobe.edu.au, wml@nwsuaf.edu.cn}\\
Code and benchmark: \url{https://github.com/Jinec98/DG-PIC}
}

\maketitle

$\blfootnote{$*$ Equal contributions.}$
$\blfootnote{\textsuperscript{\Letter} Corresponding authors.}$

\begin{abstract}
Recent point cloud understanding research suffers from performance drops on unseen data, due to the distribution shifts across different domains. While recent studies use Domain Generalization (DG) techniques to mitigate this by learning domain-invariant features, most are designed for a single task and neglect the potential of testing data. Despite In-Context Learning (ICL) showcasing multi-task learning capability, it usually relies on high-quality context-rich data and considers a single dataset, and has rarely been studied in point cloud understanding. In this paper, we introduce a novel, practical, multi-domain multi-task setting, handling multiple domains and multiple tasks within one unified model for domain generalized point cloud understanding. To this end, we propose Domain Generalized Point-In-Context Learning (DG-PIC) that boosts the generalizability across various tasks and domains at testing time. In particular, we develop dual-level source prototype estimation that considers both global-level shape contextual and local-level geometrical structures for representing source domains and a dual-level test-time feature shifting mechanism that leverages both macro-level domain semantic information and micro-level patch positional relationships to pull the target data closer to the source ones during the testing. Our DG-PIC does not require any model updates during the testing and can handle unseen domains and multiple tasks, \textit{i.e.,} point cloud reconstruction, denoising, and registration, within one unified model. We also introduce a benchmark for this new setting. Comprehensive experiments demonstrate that DG-PIC outperforms state-of-the-art techniques significantly. 
\keywords{Point Cloud Understanding \and Test-time Domain Generalization \and In-Context Learning}
\end{abstract}
\section{Introduction}
\label{sec:introduction}

\begin{figure}[htb]
    \centering
    \includegraphics[width=1\linewidth]{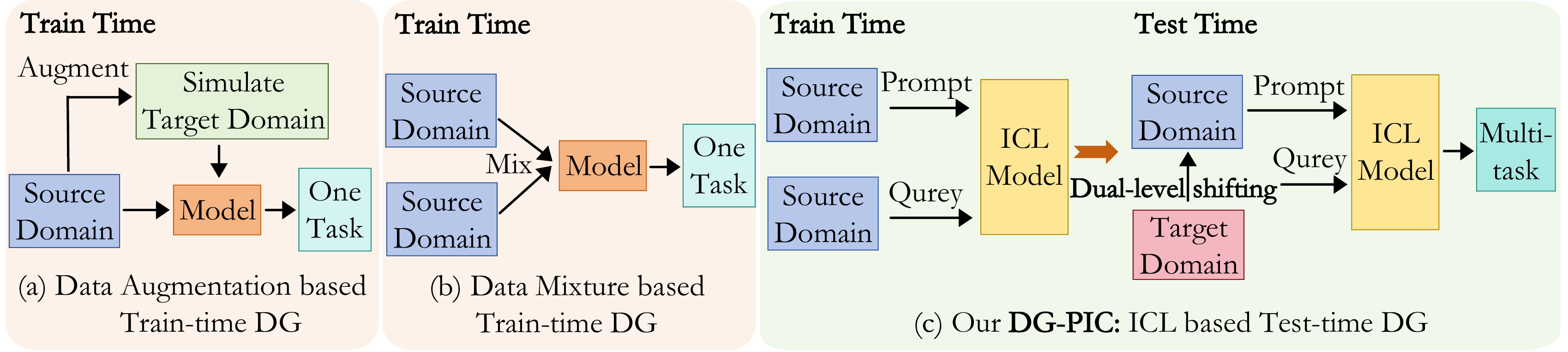}
    \caption{(a-b) Previous train-time DG techniques for point cloud understanding are typically designed for one task and dedicated to domain-invariant features at training time while ignoring the contribution of the testing data. (c) In contrast, our \textbf{DG-PIC} aims at pulling the testing features to the source ones at dual levels to improve the pre-trained model's generalizability, which does not require any model updates at testing time, excelling at the newly proposed multi-domain and multi-task setting. }
    \label{fig:teaser}
\end{figure}
 
Recent advancements in 3D point cloud understanding, pioneered by classical works like \cite{qi2017pointnet, wang2019dynamic}, have revolutionized 3D vision, facilitating applications in autonomous driving, robotics, and augmented reality. 
Despite significant development in this line of direction, current methods mainly focus on a single domain (\emph{i.e.,} training and testing on a single dataset) and have difficulties in addressing distribution shifts (\emph{i.e.,} training and testing on multiple datasets), known as the domain gap in point cloud learning. For instance, models trained on well-structured synthetic data like ModelNet40 \cite{wu20153d} may struggle to generalize to complex and noisy real-world data such as ScanObjectNN \cite{uy2019revisiting}.

Researchers have attempted to bridge this gap using Domain Adaptation (DA) techniques, with some studies \cite{yi2021complete, wang2021cross, zhao2021epointda} utilizing generative models to synthesize diverse training data to improve the model's adaptability toward target domains. Nevertheless, these methods still face challenges in handling the unseen target domains that are infeasible to acquire in the training. Thus, Domain Generalization (DG) was introduced for 3D point cloud understanding. 
These DG studies \cite{lehner20223d, sanchez2023domain, wei2022learning} concentrate on learning domain-invariant representations by training models to be more adaptable to unseen domains.  
However, as illustrated in Figure \ref{fig:teaser}, they involve two primary limitations: 
\emph{1)}  They are specifically designed for one task and lack the capacity to handle multiple tasks simultaneously. 
\emph{2)} They mainly emphasize learning domain-invariant features at training-time, while neglecting the potential of testing data as valuable resources in enhancing the generalizability, leading to less-desirable results. 

In-Context Learning (ICL) stands out as a crucial technique for multi-task learning. It trains a single model to perform multiple tasks using contextual information from input data \cite{rubin2021learning, liu2021makes, min2022rethinking}, applying the adaptive learned knowledge across various tasks and eliminating the need for distinct models for each different task. 
However, ICL has been rarely studied in point cloud understanding. Point In-Context (PIC) \cite{fang2023explore}, a specific form of ICL on 3D point clouds, aims to handle multiple point cloud tasks using a unified model. Nonetheless, PIC suffers from its dependence on input data quality and diversity to provide meaningful contexts. Also, PIC is constrained to a single dataset, potentially hindering the model's ability to generalize across different datasets and perform effectively on new unseen data deviating significantly from the trained data. 
As such, while ICL including PIC present advanced strategies for multi-task learning, they are intricately tied to the nature of the training data and have limited ability in mitigating the domain gaps between diverse datasets. 

Motivated by the above analysis, we propose a novel setting that is practical and important in the real world, \emph{i.e.,} handling multiple domains and tasks within a unified model for point cloud learning. 
To this end, we introduce an innovative framework, namely Domain Generalized Point In-Context Learning (DG-PIC), for point cloud understanding. 
In particular, we employ PIC during pre-training to gather rich, generalizable information across all source domains, creating an off-the-shelf pre-trained model, and introduce test-time DG to enhance the model's generalizability by aligning the unseen target features with source domains without any model updates during the testing.

Our DG-PIC comprises two key, novel modules. Firstly, we design a dual-level module to estimate the source prototypes, considering both global-level and local-level features. This dual-context learning strategy ensures a comprehensive understanding of the source domain, through capturing rich shape patterns and fine geometric structure details. 
Secondly, during testing time, we design the dual-level feature shifting scheme to push the target data towards the source domains, leveraging both macro-level domain-aware semantic information and micro-level patch-aware positional structures. Subsequently, we select the most similar sample from the nearest source domain to generate the query-prompt pair. 
In this manner, our DG-PIC effectively bridges the gap between the source and target domains without requiring model updates at testing time, and handles multiple tasks, within a unified model, leading to strong generalization ability and superior performance in terms of the new multi-domain and multi-task setting. 
In addition, we introduce a new benchmark for evaluating performance in this proposed setting. We meticulously select a total of $30,954$ point cloud samples from $4$ distinct datasets, including $2$ synthetic datasets (ModelNet40 \cite{wu20153d} and ShapeNet \cite{chang2015shapenet}) and $2$ real-world datasets (ScanNet \cite{dai2017scannet} and ScanObjectNN \cite{uy2019revisiting}), encompassing $7$ same object categories, and generate corresponding ground truth for $3$ different tasks (reconstruction, denoising,
and registration). 

Our main contributions are summarized as follows:
\begin{itemize}
    \item We introduce a novel and practical multi-domain multi-task setting in point cloud understanding and propose the DG-PIC, the first network to handle multiple domains and tasks within a unified model for test-time domain generalization in point cloud learning.
    \item We devise two innovative dual-level modules for DG-PIC, \emph{i.e.,} dual-level source prototype estimation that considers global-level shape and local-level geometry information for representing source domain, and the dual-level test-time target feature shifting that pushes the target data towards the source domain space with leveraging macro-level domain information and micro-level patch relationship information.
    \item We introduce a new benchmark for the proposed multi-domain and multi-task setting. Comprehensive experiments show that our DG-PIC achieves state-of-the-art performance on three different tasks.
\end{itemize}
\section{Related Work}
\label{sec:relatedwork}

\subsection{Point Cloud Understanding}
Point cloud learning has become increasingly important in 3D vision \cite{xie2020pointcontrast, jiang2023masked, jiang2024dhgcn}. Pioneering works include PointNet \cite{qi2017pointnet}, which directly processes point clouds by learning spatial encodings with pooling operation, and PointNet++ \cite{qi2017pointnet++}, which employs hierarchical processing to capture local structures at multiple scales. In addition, DGCNN \cite{wang2019dynamic} updates the graph in feature space to capture dynamic local semantic features, while PCT \cite{guo2021pct} addresses global context and dependencies within point clouds using an order-invariant attention mechanism. 
Recently, methods like Point-BERT \cite{yu2022point} and Point-MAE \cite{pang2022masked} have revolutionized point cloud learning by integrating strategies from Natural Language Processing (NLP). They introduced the Masked Point Modeling (MPM) framework as a pre-text task for reconstructing masked point clouds, with Point-BERT employing a BERT-style pre-training approach to enhance performance in downstream tasks and Point-MAE utilizing masked autoencoders for self-supervised learning to obtain rich representations without labeled data. 

\subsection{In-Context Learning}
In-Context Learning (ICL) \cite{radford2021learning, rubin2021learning,yang2024exploring,wu2024glance,li2024configure}, popular in NLP, has significantly influenced the field of deep learning, particularly with Transformer-based models like GPT \cite{brown2020language} and BERT \cite{devlin2018bert}, which enables auto-regressive language models to infer unseen tasks by conditioning inputs on specific query-prompt pairs, \emph{i.e.,} the contexts. 
The principles of ICL, primarily learning from the given data context without explicit task-specific instructions, have been extended beyond NLP, into computer vision and 3D data analysis. For instance, SupPR \cite{zhang2023makes} delved into the importance of in-context examples in enhancing the inference abilities of large-scale vision models for unseen tasks, presenting a prompt retrieval framework that automates prompt selection. 
Painter \cite{wang2023images} generalized visual ICL, facilitating the model to paint an image through various task prompts. In 3D, PIC \cite{fang2023explore} investigated the ICL paradigm for enhancing the 3D point cloud understanding, revealing the model's applicability for conducting multi-tasks concurrently. 
Though these methods showcase the generalizability across various tasks to some extent, they all focus on a single dataset (\emph{i.e.,} single domain), overlooking the disparities between different domains. In fact, it is common and practical to have different domains in point cloud learning. This essentially motivates the design of our method: a unified model for multi-domain, multi-task point cloud learning. 

\subsection{Point Cloud Adaptation and Generalization}
Domain Adaptation (DA) \cite{zhou2022generative,zhou2023context,zhou2023self,gu2021pit,zhou2022uncertainty,song2024ba,zhang2020unsupervised,zhang2024dual,xiong2023confidence,xiong2024pyra,xiao2024cat} and Domain Generalization (DG) \cite{long2024rethink,wang2024disentangle,zhou2024test,zhou2023instance,zhou2022adaptive,long2023diverse,long2024dgmamba,lu2023neuron} are introduced to enhance the adaptability and generalizability of models across diverse datasets. In point cloud analysis, DA~\cite{zou2021geometry,xu2021spg, achituve2021self, shen2022domain, saleh2019domain,liu2024cloud} focuses on transferring knowledge from a labeled source domain to an unlabeled target domain, often employing unsupervised learning techniques to tackle challenges posed by the sparse and irregular nature of point cloud data \cite{wang2023ssda3d, wu2019squeezesegv2, ding2022doda, katageri2024synergizing, yang2022no, sinha2023mensa}.  PointDAN \cite{qin2019pointdan} addressed domain shift among the local-level geometric structures, while GLRV \cite{fan2022self} presented a self-supervised learning strategy and a pseudo-labeling method for point cloud adaptation. 
Differently, DG~\cite{zhao2022synthetic, huang2023sug, qu2023modality,lu2023towards,xiao20233d} aims to train models that perform effectively across various unseen point cloud datasets, thereby enhancing generalizability 
\cite{li2023bev, wang2023towards}. SemanticSTF \cite{xiao20233d} used random data augmentation across geometry styles and applied contrastive learning to learn general features. DGLSS \cite{kim2023single} simulated the unseen domain during training and built domain-invariant scene-wise class prototypes. The methods mentioned above typically introduce known target domains directly during training or mimic them, struggling to handle diverse unseen test data. 
In contrast, test-time DG methods \cite{park2023test, zhao2022test}  shifted testing sample features to the nearest source domain without additional model updates at test-time, improving robustness and performance on previously unseen domains.
These advancements are useful in addressing the domain gap between synthetic and real-world data. However, they are usually task-specific and lack generalization capabilities on multi-task learning in which a unified network model tackles different tasks. By contrast, we consider both different domains and different tasks.

\section{Methodology}
\label{sec:method}
\begin{figure*}[htb]
    \centering
    \includegraphics[width=1\linewidth]{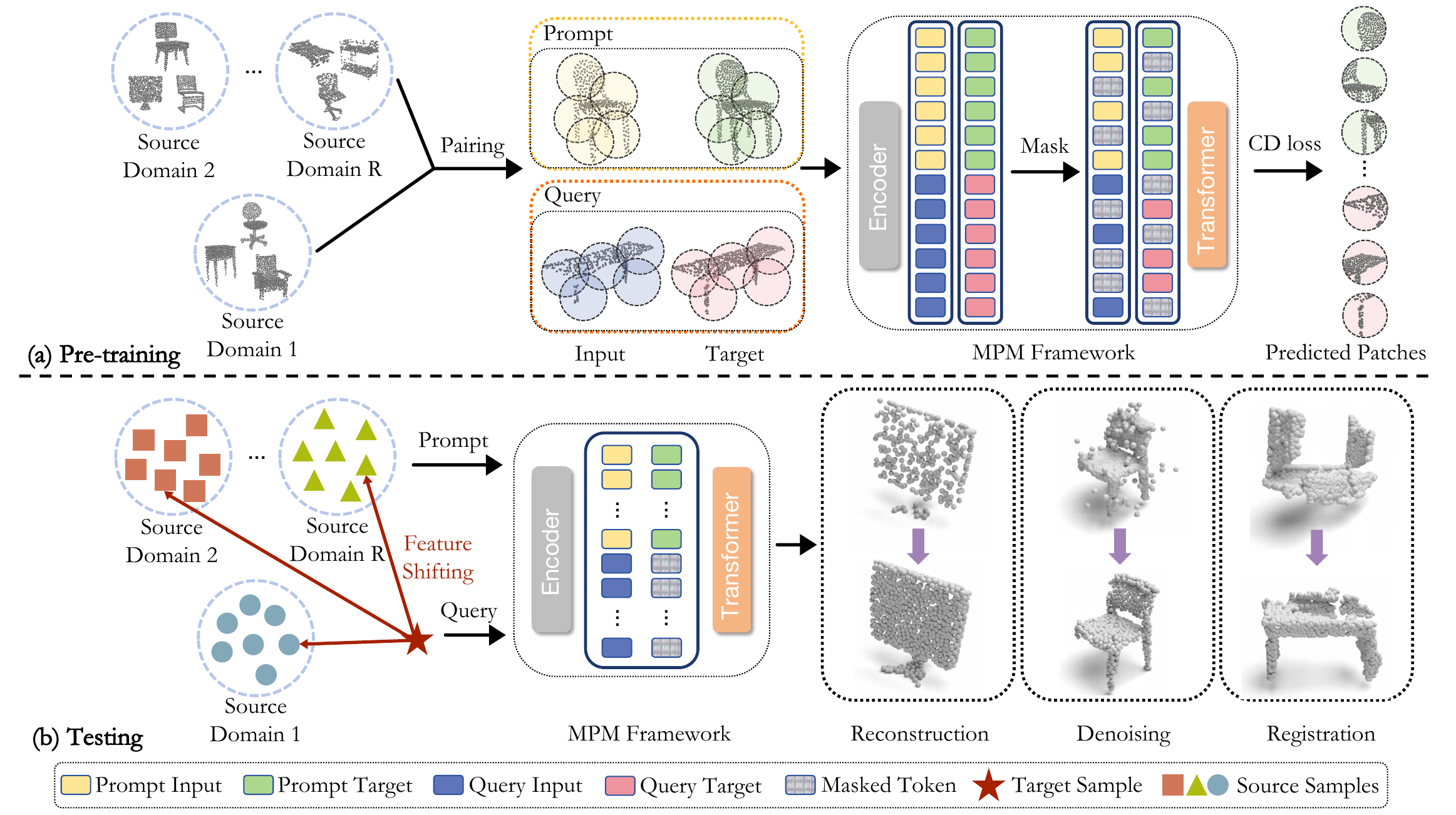}
    \caption{\textbf{The proposed DG-PIC.} (a) Pre-training: we select an arbitrary sample from source domains and form a query-prompt pair with the current one. The point cloud pairs will tackle the same task. Then, we mask some patches in the target point clouds randomly through the MPM framework and reconstruct them via the Transformer model. 
    (b) Testing: we freeze the pre-trained model, and generalize it towards unseen target sample through two key components: estimating source domain prototypes using \textit{global-level} and \textit{local-level} features, aligning target features with source domains by considering \textit{macro-level} semantic information and \textit{micro-level} positional relationships. We select the most similar sample from the nearest source as the prompt.}
    \label{fig:overview}
\end{figure*}

We propose a new multi-domain and multi-task setting for test-time domain generalized point cloud understanding aimed at attaining a universally generalized model. 
As shown in Figure \ref{fig:overview}, we design an innovative method that accommodates both Domain Generalization and Point In-Context Learning, namely \textbf{DG-PIC}. 
In DG-PIC, we devise dual-level source prototype estimation and dual-level test-time feature shifting modules, allowing the pre-trained model to generalize to unseen domain data by aligning the test data toward the source ones, without requiring any model updates at testing time.
Lastly, we introduce a new benchmark for evaluating performance in this new setting. 

\subsection{Revisiting Point In-Context (PIC) Learning Model}
PIC is a 3D point cloud In-Context Learning (ICL) procedure inspired by 2D visual ICL principles. Through the Masked Point Modeling (MPM) framework \cite{yu2022point}, point clouds are converted into a sentence-like format, which is then encoded into tokens. Tokens are then reconstructed to generate diverse task-specific outputs. 

\noindent \textbf{In-Context Learning in 3D.}
During the training, each input sample comprises two pairs of point clouds: the input point clouds and their corresponding targets addressing the same task. Following previous works \cite{yu2022point, pang2022masked}, PIC \cite{fang2023explore} employs the MPM framework, using a Transformer model for masked point cloud reconstruction.
In the inference stage, the input comprises the query point cloud and the prompt point cloud, while the output point clouds, \emph{i.e.,} the target of the given task, include the prompt target along with the query masked tokens. 
Within the MPM framework, the Transformer generates masked point cloud patches based on various task prompts, and the corresponding point cloud is derived from a unified model. Thus, PIC deduces query results for various downstream point cloud tasks, achieving multi-task objectives through a unified model. 

\noindent \textbf{Loss Function.} 
The Chamfer Distance (CD) serves as the training loss, comparing the predicted masked patch $P$ with its corresponding ground truth $G$: 
\begin{equation}
\mathrm{CD}(P, G)= \frac{1}{|P|} \sum_{x \in P} \min _{y \in G}\|x-y\|_2^2 + \frac{1}{|G|} \sum_{y \in G} \min _{x \in P}\|y-x\|_2^2.
\end{equation}

\subsection{Multi-Domain in Point In-Context Learning}
While PIC investigated the capability of a multi-task generalized model within a single domain, it remains unexplored for multiple domains. Meanwhile, significant differences exist between various data sources, like diverse data distributions and resolutions, making it challenging for the model to uniformly handle all different data, and leading to limited generalization ability. 

\noindent \textbf{Problem Setting: Test-time DG in PIC.}
Our innovation lies in test-time DG for PIC to tackle discrepancies among different domains. 
During the pre-training stage, we employ PIC to acquire rich information that is generalizable across all source domains, treating it as the off-the-shelf pre-trained model. The test-time DG is to facilitate the model's generalizability by aligning the unseen target domain $D_t$ towards source domains $D_s$ that share correlative features without any model update at testing time.
We define $\left\{I_q, I_p\right\}, \left\{T_q^k, T_p^k\right\}$ as the input and target point clouds pair, where $k$ represents the task index while $q$ and $p$ indicate the \textit{query} and \textit{prompt} sample, respectively. Suppose we have $R$ source domains $D_s = \left\{D_s^1, D_s^2, \dots, D_s^R\right\}$ that are available at pre-training stage. 
Let $\nu, \tau$ denote samples from two different domains. In practice, the feature gap $||F_\theta(\nu) - F_\theta(\tau)||$, where $F_\theta(\cdot)$ is the shared feature encoder parameterized by $\theta$, calculated by a unified model, might be large, especially when one new domain serves as the target domain, as the model is unfamiliar with the new data. Thereby, we aim to minimize performance degradation caused by domain gaps in the unified model and maximize its potential for testing target data.  

\noindent \textbf{Multi-task Setup.}
ICL requires maintaining the consistency between inputs and outputs within the same space, \emph{i.e.,} the task involves XYZ coordinate regression. Thus, following PIC~\cite{fang2023explore}, we undertake three distinct tasks on the point cloud, including:
\emph{1)} \textit{Reconstruction}, aiming at reconstructing a dense point cloud from a sparse point set.
\emph{2)} \textit{Denoising}, striving to obtain a clear object shape by removing Gaussian noise from the input point cloud.
\emph{3)} \textit{Registration}, focusing on restoring the original orientation from a randomly rotated point cloud. 

\noindent \textbf{Multi-domain Multi-task Benchmark.}
We propose a new setting, namely, the multi-domain and multi-task setting. To the best of our knowledge, there are no available benchmarks for performance evaluation. Therefore, we introduce a new benchmark wherein we meticulously curate and select data from $4$ distinct datasets (comprising $2$ synthetic and $2$ real-world datasets) with $7$ same object categories, and generate the corresponding ground truth based on $3$ different tasks. 
Specifically, for the synthetic datasets, we employ ModelNet40 \cite{wu20153d} and ShapeNet \cite{chang2015shapenet}, with ModelNet40 choosing $3,713$ samples for training and $686$ for testing, and ShapeNet selecting $15,001$ training samples and $2,145$ testing samples. As for real-world data, we use ScanNet \cite{dai2017scannet} and ScanObjectNN \cite{uy2019revisiting}. ScanNet provides annotations for individual objects in the real scan 3D scenes, we select $5,763$ samples for training and $1,677$ for testing. ScanObjectNN contains $1,577$ training samples and $392$ testing samples, and we merge `desk' and `table' in the dataset as `table'. Following  PIC, we normalize both the inputs and targets to only contain $1,024$ points with XYZ coordinates, and all data are conducted with various random operations, including rotation, scaling, and perturbation.

\begin{figure}[htb]
    \centering
    \includegraphics[width=1\linewidth]{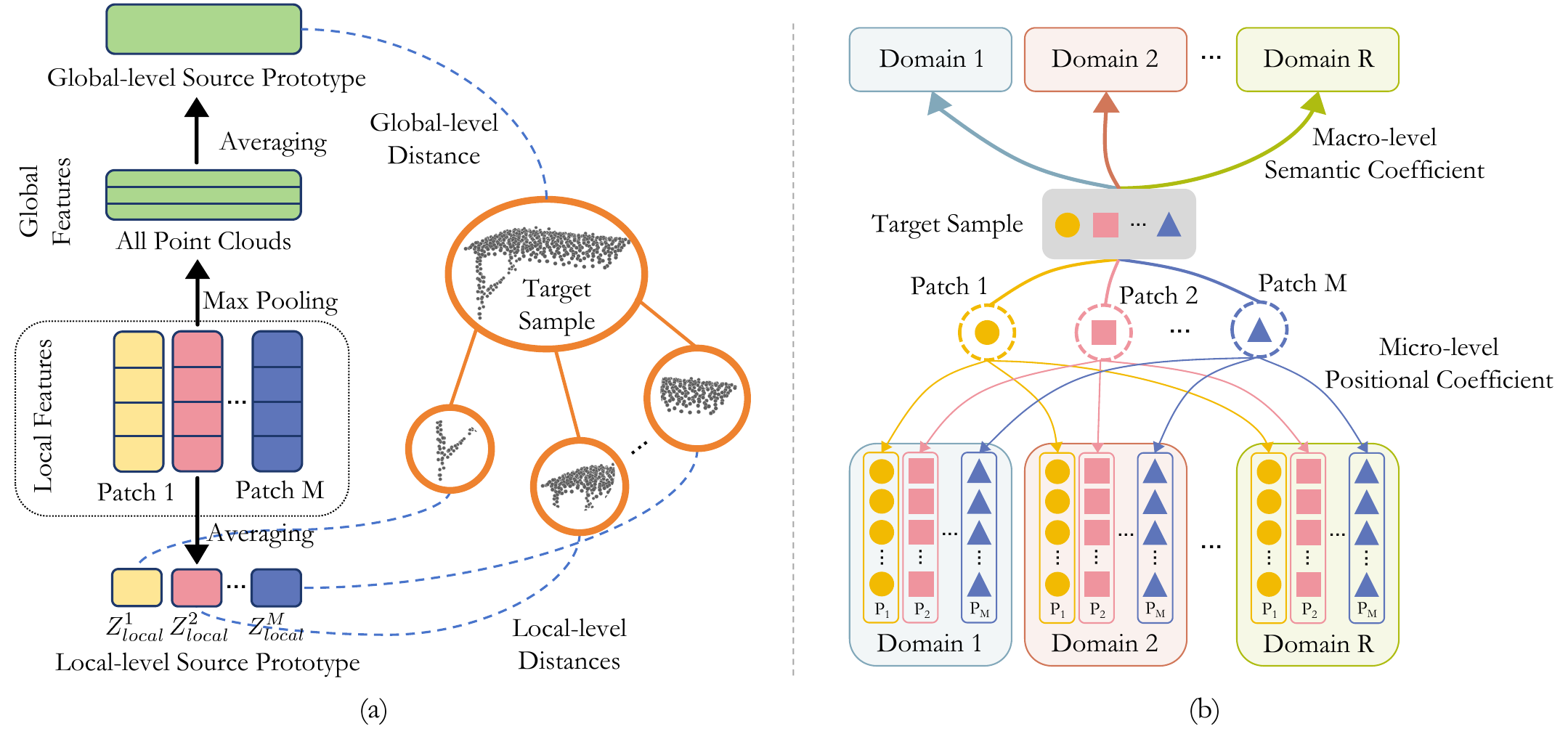}
    \caption{Illustration of the presented method: \textbf{(a) Dual-level Source Prototype Estimation.} We estimate the prototype of the source domains at the global and local levels and consider the feature distance from the target sample to the prototypes at the dual levels. \textbf{(b) Dual-level Test-time Target Feature Shifting.} We shift the target feature by considering both macro-level semantic information across all source domains and micro-level positional relationships within corresponding patches. }
    \label{fig:prototype}
\end{figure}

\subsection{Framework of DG-PIC}
Test-time DG should attempt to mitigate the domain gap and shift the testing data at the testing time to boost the generalizability. Given that 3D point clouds mainly have global shape information and local geometric structures, \emph{i.e.,} dual-level features, we propose a novel \textit{dual-level source-domain prototype estimation} module along with a \textit{dual-level target feature shifting} strategy, to enable the network to generalize to previously unseen target domains.

\noindent  \textbf{Multi-domain Prompt Pairing.}
During the training of PIC, the process begins with selecting another point cloud pair to serve as a prompt, guiding the model in performing a specific task. Both the prompt point cloud pair and the query point cloud pair are required to execute the same task. Unlike PIC which selects a prompt from a single training set (\emph{i.e.,} single domain), we randomly collect samples from various source domains, reinforcing the correlation between these domains. Let  $Trans(\cdot)$ denote as the Transformer blocks in the MPM framework, and query point clouds belonging to $D_s^i$ as $\left\{I_i, T_i^k\right\}$ and prompt point clouds in other domain $D_s^j(j\neq i)$ as $\left\{I_j, T_j^k\right\}$. Then the predicted masked patch $P$ can be depicted as: 
\begin{equation}
    P\sim(D_s^i, D_s^j) = Trans([F_\theta(I_i) \oplus F_\theta(T_i^k) \oplus F_\theta(I_j) \oplus F_\theta(T_j^k)], Mask),
\end{equation}
where $\oplus$ represents the concatenate operation that merges all point cloud tokens, and $Mask$ denotes the masked token utilized to replace the invisible token. 

\noindent \textbf{Dual-level Source Prototype Estimation.}
As for test-time DG, it is essential to estimate the prototype $Z$ for each source domain. For this estimation, we propose to respectively average global and local features across all data within the domain, as shown in Figure \ref{fig:prototype} (a). Specifically, we treat the patch-wise tokens $\left\{F_\theta(P_1), F_\theta(P_2), \dots, F_\theta(P_M)\right\}$ as the local features of each patch $P_m$ ($m\in[1, M]$, where $M$ is the patch number), which can be derived from the MPM encoder. Thereby, the \textit{local-level} prototype $Z_{local}\in\mathbb{R}^{C\times M}$ of $D_s^i$ is formulated as:
\begin{equation}
    Z_{local}^{i,m} = \frac{1}{N_{D_s^i}} \sum_{n=1}^{N_{D_s^i}} F_\theta(P_m),\quad m\in[1, M],
\end{equation}
where $N_{D_s^i}$ denotes the sample number of $D_s^i$. 
Following previous work \cite{qi2017pointnet}, we use the pooling operation to extract the global feature of the point cloud, and we can obtain the \textit{global-level} prototype $Z_{global}^i \in \mathbb{R}^C$ as:
\begin{equation}
    Z_{global}^i = \frac{1}{N_{D_s^i}} \sum_{n=1}^{N_{D_s^i}} max(F_\theta(P_m)),\quad m\in[1, M],
\end{equation}
where $max$ indicates the max pooling over the point number channel. We need to pre-train the model under the multi-domain pairing settings and store all source prototypes from the final epoch. In this way, we regard the source prototypes as the shared common information available to the target domain during testing.

Given a test sample $I_t$ in the target domain $D_t$,  we can obtain its local features $\left\{F_{local}^m \mid F_\theta(P_m), m\in[1, M]\right\}$ along with its global feature $F_{global}$ through the max pooling. We then calculate the feature Euclidean  distances $\mathcal{E}$ between the test sample and each source domain prototype at both global and local levels:
\begin{equation}
\label{equ:distance}
\begin{split}
    \mathcal{E}_{global}^i &= ||F_{global} - Z_{global}^i||,\quad i \in [1, R],\\
    \mathcal{E}_{local}^{i,m} &= || F_{local}^m - Z_{local}^{i,m}||,\quad i \in [1, R], m \in [1,M].
\end{split}
\end{equation}

Then, we retrieve the similarities between each source domain and the testing sample by comparing the global-local feature distances and consider their prototypes $Z$ as anchors to align the testing features closer to the source domains. 

\noindent \textbf{Dual-level Test-time Target Feature Shifting.}
We introduce a \textit{macro-level} semantic coefficient $\alpha$, derived from the domain-aware shape information, \emph{i.e.,} the source domain feature similarities $\mathcal{E}_{global}$, to regulate the impact from various source domains on feature shifting. 
Since the MPM framework models point cloud with tokens (\emph{i.e.,} at the patch level), the shifting can be depicted as:
\begin{equation}
\label{equ:alpha}
\begin{split}
    \alpha = softmax(\mathcal{E}_{global}), \quad
    F_{local}' = \frac{1}{R} \sum_{i=1}^R (\alpha_i F_{local} + (1-\alpha_i) Z_{local}^i).
\end{split}
\end{equation}

To consider the \textit{micro-level} geometric structure of the point cloud, as illustrated in Figure \ref{fig:prototype} (b), we further propose a patch-aware positional coefficient $\beta$ to investigate the feature relationships within point cloud patches from different domains at each patch position in 3D space. 
If two point clouds have high semantic similarity, their corresponding patches occupying the same location should hold highly similar geometrical structures, even from different domains. 
For instance, in table point clouds derived from either real scans or synthetic data, the outer regions typically feature sharper edge structures, whereas the interior regions consist of surface planes.
The designed positional coefficient is dedicated to aligning the test sample patches towards the source domains located at the same position, aiming to mitigate the discrepancies between them. Based upon this, we design the test sample feature shifting as:
\begin{equation}
\label{equ:beta}
\begin{split}
    \beta^i = softmax(\mathcal{E}_{local}^i), \quad
    F_{local}' &= \frac{1}{R} \sum_{i=1}^R \alpha_i ( \frac{1}{M}\sum_{m=1}^M \beta^{i,m} F_{local}^m)\\
    &+ \frac{1}{R} \sum_{i=1}^R (1-\alpha_i) (\frac{1}{M} \sum_{m=1}^M (1-\beta^{i,m}) Z_{local}^{i,m}).
\end{split}
\end{equation}

Through shifting, the target data is pulled closer to source domains, facilitating the model to achieve comparable performance to that observed on sources. 

\noindent \textbf{Test-time Prompt Selection.}
Upon computing the feature distances at both global and local levels between the test sample and each source domain, we can simply identify the closest source domain $D_s^t$ by:
\begin{equation}
\label{equ:lambda}
    \mathcal{E}^i = \lambda \cdot \mathcal{E}_{global}^i + (1-\lambda) \cdot \frac{1}{M} \sum_{m=1}^M \mathcal{E}_{local}^{i,m},
\end{equation}
where $\lambda$, default to $0.5$, serves as a balancing factor for adjusting the two parts in the equation. 
Therefore, the most similar sample $\left\{I_s, T_s^k\right\}$ within this source domain can be easily identified through feature distance. This similarity usually manifests similar global shapes or local geometrical structures. We select this sample as the prompt for the current test data, and thus the patch (forms the query output) predicted by the MPM framework can be represented by:
\begin{equation}
    P\sim(I_t, D_s^t) = Trans(F_{local}' \oplus Mask \oplus F_\theta(I_s) \oplus F_\theta(T_s^k)).
\end{equation}

In this manner, the divergence between the source-target domain can be mitigated by leveraging the source prototype at test-time, considering both the point cloud low-level feature and high-level task aspects. Our DG-PIC is a unified multi-domain and multi-task model for point cloud learning. In contrast, existing methods rely solely on source data or focus on a single task, often demonstrating poor generalization in this practical multi-domain and multi-task setting.
\section{Experiments}
\label{sec:experiments}

\subsection{Experimental Setting}
\label{sec:setting}
We implement our method using PyTorch and conduct experiments on two TITAN RTX GPUs. Following PIC \cite{fang2023explore}, we set the training batch size to $128$ and utilize AdamW optimizer \cite{loshchilov2019decoupled}. The learning rate is set to $0.001$, with the cosine learning scheduler and a weight decay of $0.05$. All models are trained for $300$ epochs. We sample each point cloud to $1,024$ points and split it into $G=64$ point patches, and each patch has $32$ points. We set the mask ratio to $0.7$, which is similar to previous works \cite{yu2022point, pang2022masked}. 

\subsection{Baseline and Comparison Methods}
\label{sec:compared_methods}
Since there exists no prior work on point cloud understanding in terms of multi-domain multi-task learning, we reproduce some state-of-the-art point cloud learning methods and DG techniques based on the following three schemes.  

\noindent \textbf{In-Context Learning Models.}
While ICL \cite{bar2022visual, rubin2021learning} handles multi-task learning, it lacks the capability to tackle multi-domain generalization. In comparison, we compile the source domains in our benchmark into an extensive dataset, serving as the training set for PIC. Subsequently, we transfer the trained model to the target domain for inference, thus realizing multi-domain learning. Similar to PIC \cite{fang2023explore}, we consider the baseline methods that utilize the prompt target point cloud as its prediction and the same training settings as PIC. 

\noindent \textbf{Task-specific Models.}
We select $7$ representative methods, PointNet \cite{qi2017pointnet}, PointNet++ \cite{qi2017pointnet++}, DGCNN \cite{wang2019dynamic}, PCT \cite{guo2021pct}, Point-MAE \cite{pang2022masked}, Pointmixup \cite{chen2020pointmixup}, and PointCutMix \cite{zhang2022pointcutmix} and implement them for multi-task multi-domain learning. In this setting, these methods share a backbone network while having independent task-specific heads designed for different tasks. Aligning with our method, we replicate the augmentation-based methods Pointmixup and PointCutMix as a DG model, where each sample is mixed with another random source domain. 

\noindent \textbf{Multi-task Models.}
For a fair comparison, we also designed a multi-task model for the $5$ above-mentioned methods, \emph{i.e.,} utilizing a shared backbone network and head to simultaneously learn all $3$ tasks. 

\subsection{Main Results}
\label{sec:results}
\begin{figure*}[tbp]
    \centering
    \includegraphics[width=1\linewidth]{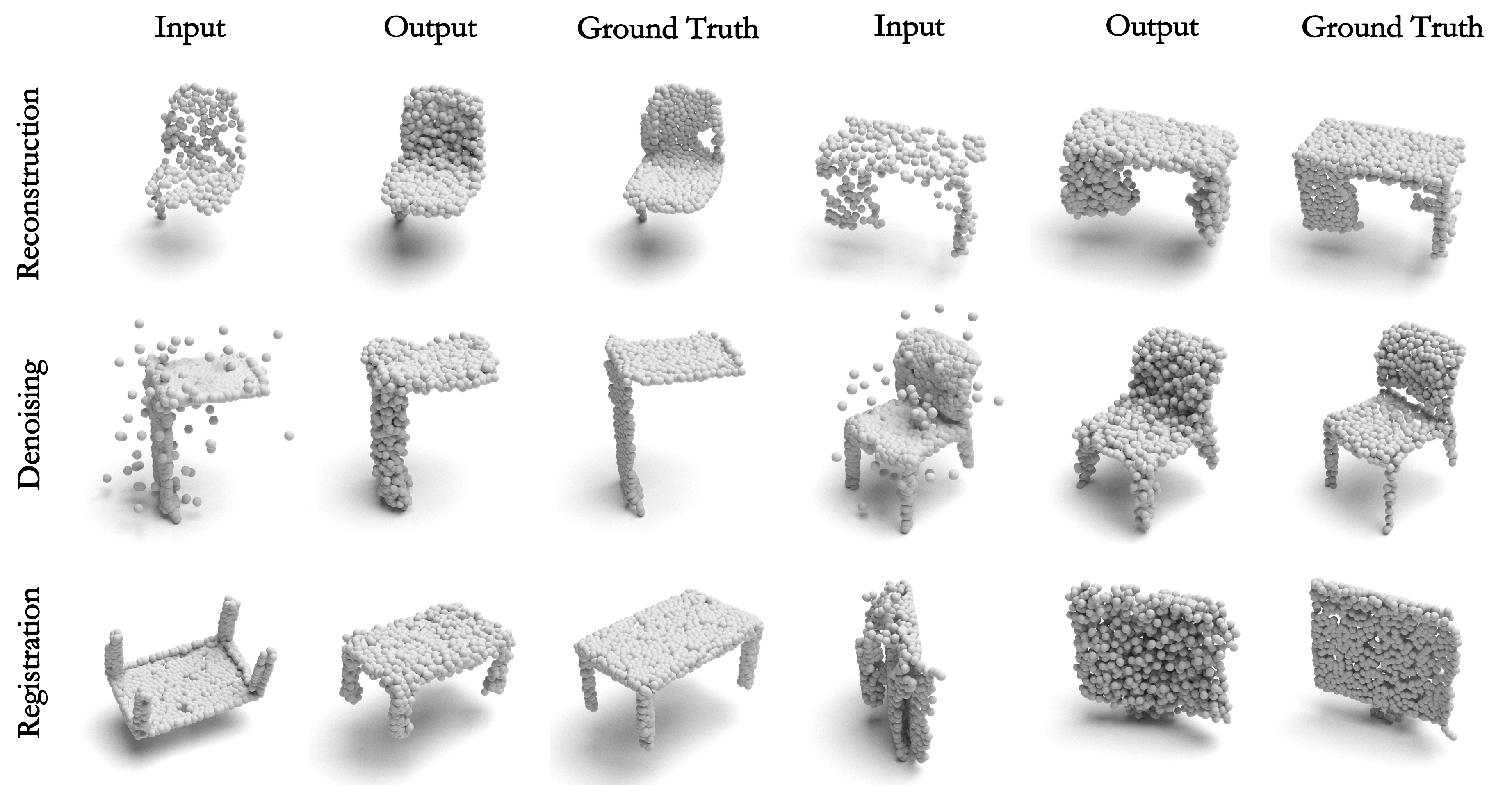}
    \caption{\textbf{Visualization results} of our DG-PIC and their corresponding targets (denoted as ground truth) under $3$ different tasks, including reconstruction, denoising, and registration. The real-world dataset ScanObjectNN serves as the target domain. }
    \label{fig:visual_results}
\end{figure*}

Table \ref{tab:compared_method} presents comprehensive experimental results for our DG-PIC model and other methods across diverse tasks on the novel benchmark we introduced, including reconstruction, denoising, and registration. As we can see, our method demonstrates remarkable performance on different tasks, outperforming all compared methods. Moreover, multi-domain also brings new challenges to the model's generalization ability, but our unified model can fulfill this without requiring additional training. As shown in Figure \ref{fig:visual_results}, we visualize the results of our DG-PIC model, from which we can find that our model can generate quality predictions of the unseen target domain by employing distinct prompts from its closest source domain among $3$ different tasks. In this section, we adopt the ScanObjectNN dataset as the target domain unless otherwise stated. 

\begin{table*}[tbp]
    \caption{Comparison of three training schemes with three point cloud tasks on our benchmark. Each model is trained on other source domains and tested on the target domain ScanObjectNN. `Fully Supervised': the model is supervised by the corresponding ground truth without incorporating any generalization strategy. We use the Chamfer Distance ($\times 10^{-3}$) as the metric. (The lower value indicates a better performance).  }
    \centering
    \resizebox{0.97\textwidth}{!}{
    \begin{tabular}{lccccc}
    \hline
         Methods & Venue & Setting & Recon. & Denoi. & Regis.\\
    \hline
         \multicolumn{6}{c}{Task-specific Models}\\
    \hline
         PointNet \cite{qi2017pointnet} & CVPR 2017 & Fully Supervised Learning & 41.1 & 41.9 & 43.5\\
         PointNet++ \cite{qi2017pointnet++} & NeurIPS 2017 & Fully Supervised Learning & 40.3 & 38.5 & 41.9\\
         DGCNN \cite{wang2019dynamic} & TOG 2019 & Fully Supervised Learning & 39.0 & 37.9 & 39.8\\
         PCT \cite{guo2021pct} & CVM 2021 & Fully Supervised Learning & 30.8 & 35.8 & 32.5\\
         Point-MAE \cite{pang2022masked} & ECCV 2022 & Fully Supervised Learning & 30.4 & 36.0 & 31.2\\
         Pointmixup \cite{chen2020pointmixup} & ECCV 2020 & Train-time Domain Generalization & 40.5 & 42.3 & 41.7\\
         PointCutMix \cite{zhang2022pointcutmix} & Neuro. 2022 & Train-time Domain Generalization & 44.8 & 43.5 & 41.3\\
    \hline
         \multicolumn{6}{c}{Multi-task Models}\\
    \hline
         PointNet \cite{qi2017pointnet} & CVPR 2017 &  Fully Supervised Learning & 41.3 & 43.6 & 45.6\\
         PointNet++ \cite{qi2017pointnet++} & NeurIPS 2017 & Fully Supervised Learning & 40.9 & 39.6 & 43.2\\
         DGCNN \cite{wang2019dynamic} & TOG 2019 & Fully Supervised Learning & 40.7 & 38.2& 41.6\\
         PCT \cite{guo2021pct} & CVM 2021 & Fully Supervised Learning & 31.5 & 36.5 & 34.9\\
         Point-MAE \cite{pang2022masked} & ECCV 2022 & Fully Supervised Learning & 30.7 & 36.3 & 31.0\\
         Pointmixup \cite{chen2020pointmixup} & ECCV 2020 & Train-time Domain Generalization & 40.9 & 44.2 & 43.0\\
         PointCutMix \cite{zhang2022pointcutmix} & Neuro. 2022 & Train-time Domain Generalization & 45.9 & 47.5 & 43.6\\
    \hline
         \multicolumn{6}{c}{In-Context Learning Models}\\
    \hline
         Baseline & & Fully Supervised Learning & 156.2 & 127.1 & 83.6\\
         PIC \cite{fang2023explore} & NeurIPS 2023 & Fully Supervised Learning & 72.9 & 80.0 & 12.7\\
         Our \textbf{DG-PIC} & ECCV 2024 & Test-time Domain Generalization & \textbf{4.1} & \textbf{15.2} & \textbf{5.8}\\
    \hline
    \end{tabular}
    }
    \label{tab:compared_method}
\end{table*}

\noindent \textbf{Comparison to Conventional Methods.} 
Several conventional point cloud learning methods, such as PointNet \cite{qi2017pointnet}, DGCNN \cite{wang2019dynamic}, and PCT \cite{guo2021pct}, often struggle with achieving generalization. Despite incorporating individual heads for these methods to suit diverse tasks (\emph{i.e.,} task-specific scheme), our method consistently outperforms them across all tasks. They exhibit inferior performance with a shared network in a multi-task scheme. Our method is characterized as a fully unified model, capable of overcoming domain gaps through the multi-task in-context learning setting. Notably, these selected models suffer from significant performance drops when applying to the new dataset, demonstrating the superiority and better generalization capacity of our method. This is mainly due to our proposed test-time DG feature shifting module and ICL strategy.

\noindent \textbf{Comparison to DG Methods.}
We compare our DGPIC with a widely-used DG strategy, specifically the mix strategy. We adopt representative works Pointmixup \cite{chen2020pointmixup} and PointCutMix \cite{zhang2022pointcutmix}, by employing mixing data from various source domains during training. 
However, as shown in Table \ref{tab:compared_method}, these methods usually explore generalization between domains while overlooking task-specific differences, as indicated by the consistently high Chamfer Distance results with minimal variance across all tasks. This shows the limitation of relying solely on DG and proves insufficient for addressing multi-domain multi-task scenarios. 

\noindent \textbf{Comparison to ICL Method.} 
ICL methods such as PIC \cite{fang2023explore} are effective for multi-task scenarios using a unified model but struggle with generalization across different domains. They perform poorly when faced with unseen data, failing to perform the specified task with the provided prompt in existing source domains. 
In contrast, our proposed test-time DG module effectively addresses this difficulty by aligning the test data features with familiar source domains. In addition, utilizing the most similar source sample as the prompt pair effectively mitigates the gap between the source and target. This enables our unified model to enjoy multi-domain generalizability. 

\noindent \textit{Please refer to the supplementary for more results.}

\subsection{Ablation Studies}
\label{sec:ablation}
Our two novel dual-level modules, \emph{i.e.,} global-local source prototype estimation and macro-micro target feature shifting, are pivotal for aligning the test data feature toward the source domains. To verify the effectiveness of these components, we conduct the ablation studies, and the results are shown in Table \ref{tab:ab_prototype}. 

\noindent 
\textbf{Prototype Estimation: Global-Local Levels.}
We consider both global and local prototypes when calculating the distance from the target sample to the source domain, allowing our model to understand sample data comprehensively by considering both global shape information and local geometric structure. 
We first take the prototype estimation out of our dual-level design by adopting a straightforward averaged shifting approach to align the target feature towards only one domain (Models A-D). 
We determine the anchor source by measuring the feature distance from the target sample to each source domain prototype.
An intuitive alternative would involve randomly selecting the anchor domain. As depicted in Table \ref{tab:ab_prototype} (Model A), the random feature shift may result in ambiguous features, leading to the loss of significant features inherent in the test sample and consequently yielding inferior results. 
We also compare results using only the global or local features and find that incorporating both global and local levels captures richer features, leading to superior results.

\noindent 
\textbf{Anchor Source Domains: One or All?}
We observe that pushing the target sample towards only one nearest anchor source domain may result in less abundant source domain information within the model. Consequently, we delve into the superior option of shifting target samples towards all source domains to comprehensively maximize the model's generalizability. 
As shown in Table \ref{tab:ab_prototype} (Models D and E), aligning target samples to all source domains leads to only a marginal improvement.  We attribute this to the fact that equally pulling target sample features towards all prototypes tends to flatten features across all test data, potentially leading to less discriminative features for the test sample. 
By contrast, our dual-level target feature shifting mechanism enables the model to mimic prior information about new test samples, achieving much better results. 

\noindent
\textbf{Feature Shifting: Macro-Micro Levels.}
To distinguish the degree of contributions from different source domains in feature shifting, as shown in Eq. (\ref{equ:beta}), we implicitly utilize the feature similarity between the target samples and the global-local prototype of each source domain, allowing the model to derive informative coefficients that facilitate shifting.
The comparison in Table \ref{tab:ab_prototype} (Models E-G and ours) manifests that incorporating semantic information among all domains at the macro-level and considering the positional relationships of the point cloud patches at the micro-level both contribute to the final model,  assisting in alleviating the domain discrepancy between sources and targets.

\begin{table}[htb]
    \caption{Ablations on dual-level prototype estimation and test-time feature shifting. }
    \centering
    \begin{tabular}{lrclcccc}
    \hline
         Models & Prototype & $\rightarrow$ & Shifting & Anchors & Recon. & Denoi.& Regis.\\
    \hline
         Model A & random & $\rightarrow$ & average & one domain & 8.4 & 40.5 & 6.7\\
         Model B & only global & $\rightarrow$ & average & one domain & 7.2 & 38.3 & 6.4\\
         Model C & only local & $\rightarrow$ & average & one domain & 7.3 & 36.7 & 6.7 \\
         \hdashline
         Model D & global-local & $\rightarrow$ & average & one domain & 6.8 & 35.1 & 6.2\\
         Model E & global-local & $\rightarrow$ & average & all domains & 6.3 & 32.4 & 6.5\\
         \hdashline
         Model F & global-local & $\rightarrow$ & only macro & all domains & 5.2 & 22.7 & 6.0 \\
         Model G & global-local & $\rightarrow$ & only micro & all domains & 4.9 & 25.6 & 6.2\\
    \hline
         \textbf{Our choice} & global-local & $\rightarrow$ & macro-micro & all domains & \textbf{4.1} & \textbf{15.2} & \textbf{5.8}\\
    \hline
    \end{tabular}
    \label{tab:ab_prototype}
\end{table}
\section{Conclusion}
\label{sec:conclusion} 
In this paper, we introduce the multi-domain multi-task setting that handles multiple domains and tasks within a unified model. We propose Domain Generalized Point-In-Context (DG-PIC), boosting the generalizability across different domains and tasks. Specifically, we devise dual-level source prototype estimation, capturing global and local-level features to comprehensively represent source domains, and a dual-level test-time feature shifting strategy that aligns target data with source domains at both macro and micro-level during testing. Extensive experiments demonstrate that DG-PIC outperforms state-of-the-art techniques.

\section*{Acknowledgments}
Jincen Jiang is supported by the China Scholarship Council (Grant Number 202306300023), and the Research and Development Fund of Bournemouth University. Meili Wang is supported by the National Key Research and Development Program of China (Grant Number 2022YFD1300201), 

\bibliographystyle{splncs04}
\bibliography{main}

\end{document}